\documentclass[runningheads]{llncs}
\usepackage{amsmath}
\usepackage{amssymb}
\usepackage{graphicx}
\usepackage{subcaption}
\usepackage{hyperref}
\hypersetup{hidelinks,
backref=false,
pagebackref=false,
hyperindex=false,
breaklinks=false,
colorlinks=true,
urlcolor=blue,
bookmarks=false,
bookmarksopen=false,
pdftitle={Title},
pdfauthor={Author}}

\begin{document}
\title{Post-synaptic~potential~regularization has~potential}
%
%
\author{Enzo~Tartaglione\inst{1}\orcidID{0000-0003-4274-8298} \and
Daniele~Perlo\inst{1} \and
Marco~Grangetto\inst{1}\orcidID{0000-0002-2709-7864	}}
\authorrunning{Tartaglione et al.}
%
\institute{Universit\`a degli studi di Torino, Turin, Italy}
\maketitle              
\begin{abstract}
Improving generalization is one of the main challenges for training deep neural networks on classification tasks. In particular, a number of techniques have been proposed, aiming to boost the performance on unseen data: from standard data augmentation techniques to the $\ell_2$ regularization, dropout, batch normalization, entropy-driven SGD and many more.\\
In this work we propose an elegant, simple and principled approach: post-synaptic potential regularization (PSP). We tested this regularization on a number of different state-of-the-art scenarios. Empirical results show that PSP achieves performances comparable to more sophisticated learning strategies in the MNIST scenario, while improves the generalization compared to $\ell_2$ regularization in deep architectures trained on CIFAR-10.

\end{abstract}
%
%
%



%
%

\section{Introduction}
In the last few years artificial neural network (ANN) models received huge interest from the research community. In particular, their potential capability of solving complex tasks with extremely simple training strategies has been the initial spark, while convolutional neural networks (CNNs), capable of self-extracting relevant features from images, have been the fuel for the burning flame which is the research around ANNs. Furthermore, thanks to the ever-increasing computational capability of machines with the introduction of GPUs (and, recently, TPUs) in the simulation of neural networks, ANNs might be embedded in many portable devices and, potentially, used in everyday life.\\
State-of-the-art ANNs are able to learn very complex classification tasks: from the nowadays outdated MNIST~\cite{lecun1998gradient}, moving to CIFAR-10 and then even the ImageNet classification problem. In order to overcome the complexity of these learning tasks, extremely complex architectures have been proposed: some examples are VGG~\cite{simonyan2014very} and ResNet~\cite{he2016deep}. However, due to their extremely high number of parameters, these models are prone to over-fitting the data; hence, they are not able to generalize as they should. In this case, the simple learning strategies (like SGD) are no longer able to guarantee the network to learn the relevant features from the training set and other strategies need to be adopted.\\
In order to improve the generalization of ANNs, several approaches have been proposed. One of the most typical relies on the introduction of a ``regularization'' term, whose aim is to add an extra constraint to the overall objective function to be minimized. Recently, other approaches have been proposed: from the introduction of different optimizers~\cite{kingma2014adam} to data augmentation techniques, the proposal of new techniques like dropout~\cite{srivastava2014dropout} and even changing the basic architecture of the ANN~\cite{he2016deep}.\\
In this work, we propose a regularization term inspired by a side effect of the $\ell_2$ regularization (also known ad \emph{weight decay}) on the parameters. In particular, we are going to show that, naturally, weight decay makes the post-synaptic potential dropping to zero in ANN models. From this observation, a post-synaptic potential regularization (PSP) is here proposed. Differently from $\ell_2$ regularization, its effect on the parameters is not local: parameters belonging to layers closer to the input feel the effect of the regularization on the forward layers. Hence, this regularization is aware of the configuration of the whole network and tunes the parameters using a global information. We show that the standard $\ell_2$ regularization is a special case of the proposed regularizer as well. Empirically we show that, when compared to the standard weight decay regularization, PSP generalizes better.\\
The rest of this paper is organized as follows. In Sec.~\ref{sec::relatedwork} we review some of the most relevant regularization techniques aiming at improving generalization. Next, in Sec.~\ref{sec::psp} we introduce our proposed regularization, starting from some simple considerations on the effect of the $\ell_2$ regularization on the post synaptic potential and analyzing the potential effects on the learning dynamics. Then, in Sec.~\ref{sec::exp} we show some empirical results and some extra insights of the proposed regularization. Finally, in Sec.~\ref{sec::conclusion} the conclusions are drawn.

\section{Related work}
\label{sec::relatedwork}
Regularization is one of the key features a learning algorithm should particularly take care of in deep learning, in order to prevent data over-fit and boosting the generalization~\cite{goodfellow2016deep}. Even though such a concept is more general and older than the first ANN models~\cite{tikhonov1943stability}, we are going to focus on what regularization for deep architectures (trained on finite datasets via supervised learning) is. We can divide the regularization strategies in our context under four main categories~\cite{kuko}: 
\begin{itemize}
    \item regularization via \emph{data}: some examples include (but are not limited to) the introduction of gaussian noise to the input~\cite{goldberg1998regression}, dropout to the input~\cite{srivastava2014dropout}, data augmentation~\cite{calimeri2017biomedical}~\cite{cui2015data}, batch normalization~\cite{ioffe2015batch}.
    \item regularization via \emph{network architecture}: in this case, the architecture is properly selected in order to fit the particular dataset we are aiming to train. It can involve the choice of single layers (pooling, convolutional~\cite{lecun1998gradient}, dropout~\cite{srivastava2014dropout}), it can involve the insertion of entire blocks (residual blocks~\cite{he2016deep}), the entire structure can be designed on-purpose~\cite{elsken2018neural}~\cite{caruana1997multitask}~\cite{gulcehre2016mollifying} or even pruned~\cite{weigend1991back}~\cite{tartaglione2018learning}.
    \item regularization via \emph{optimization}: an optimizer can determine the nature of the local minima and avoid ``bad'' local minima (if any~\cite{kawaguchi2016deep}), boosting the generalization~\cite{kingma2014adam}~\cite{chaudhari2016entropy}. The initialization also seemed to cover an important role~\cite{glorot2010understanding}, together with cross-validation based techniques like early-stopping~\cite{prechelt1998automatic}.
    \item regularization via \emph{regularization term}: here a regularization term is added to the loss, and a global objective function (sum of loss and regularization term \eqref{eq::J} ) is minimized. This is the scope of our work.
\end{itemize}
One of the ground-breaking regularization techniques, proposed few years ago, is \emph{dropout}. Srivava~\emph{et~al.}~\cite{srivastava2014dropout} proposed, during the training process, to stochastically set a part of the activations in an ANN to zero according to an a-priori set dropout probability. Empirically it was observed that, applying dropout on fully-connected architecture, was significantly improving the generalization, while its effectiveness was less evident in convolutional architectures. Such a technique, however, typically requires a longer training time, and sometimes a proper choice of the dropout probability may change the effectiveness of the technique. However, dropout has many variants aiming to the same goal: one of the most used is dropconnect by Wan~\emph{et~al.}~\cite{wan2013regularization}.\\
A completely different approach to boost the generalization is to focus the attention on some regions of the loss functions. It has been suggested by Lin~\emph{et~al.}~\cite{lin2017does} that ``sharp'' minima of the loss function does not generalize as well as ``wide'' minima. According to this, the design of an optimizer which does not remain stuck in sharp minima helps in the generalization. Towards this end, some optimizers like SGD or Adam~\cite{kingma2014adam} are already implicitly looking for these kind of solutions. Recently, a specially-purposed optimizer, \emph{Entropy-SGD}, designed by Chaudhari~\emph{et~al.}~\cite{chaudhari2016entropy}, showed good generalization results. However, more sophisticated optimizers increase the computational complexity significantly.\\
A regularization technique, proposed about 30 years ago by Weigend~\emph{et~al.} and just recently re-discovered, is \emph{weight elimination}~\cite{weigend1991back}: a penalty term is added to the loss function and the total objective function is minimized. The aim of the regularization term is here to estimate the ``complexity'' of the model, which is minimized together with the loss function. The learning complexity for an object increases with its number of parameters: there should exist an optimal number of parameters (or, in other words, configuration) for any given classification problem. Supporting this view, while using their \emph{sensitivity-driven regularization}~\cite{tartaglione2018learning} aiming to sparsify deep models, Tartaglione~\emph{et~al.} observed an improvement of the generalization for low compression ratios.\\
Any of the proposed regularization techniques, however, is typically used jointly to the $\ell_2$ regularization. Such a technique is broadly used during most of the ANN trainings and, despite its simple formulation, under a wide range of different scenarios, it improves the generalization. Furthermore, many recent works suggest that there is a correspondence between $\ell_2$ regularization and other techniques: for example, Wager~\emph{et~al.}~\cite{wager2013dropout} showed an equivalence between dropout and weight decay. Is there something else to understand about $\ell_2$ regularization? What's under the hood? This will be our starting point, to be discussed more in details in Sec.~\ref{ssec::L2}.

\section{Post-synaptic potential regularization}
\label{sec::psp}
In this section we first analyze the effect of weight decay on the output of any neuron in an ANN model. We show that $\ell_2$ regularization makes the post-synaptic potential drop to zero. Hence, a regularization over the post-synaptic potential is formulated (PSP). Next, the parameters update term is derived and some considerations for multi-layer architectures are drawn. Finally, we show the concrete effect of post-synaptic potential regularization on both the output of a single neuron and its parameters.

\subsection{Notation}
\begin{center}
\begin{figure}
    \centering
    \includegraphics[width=0.75\textwidth]{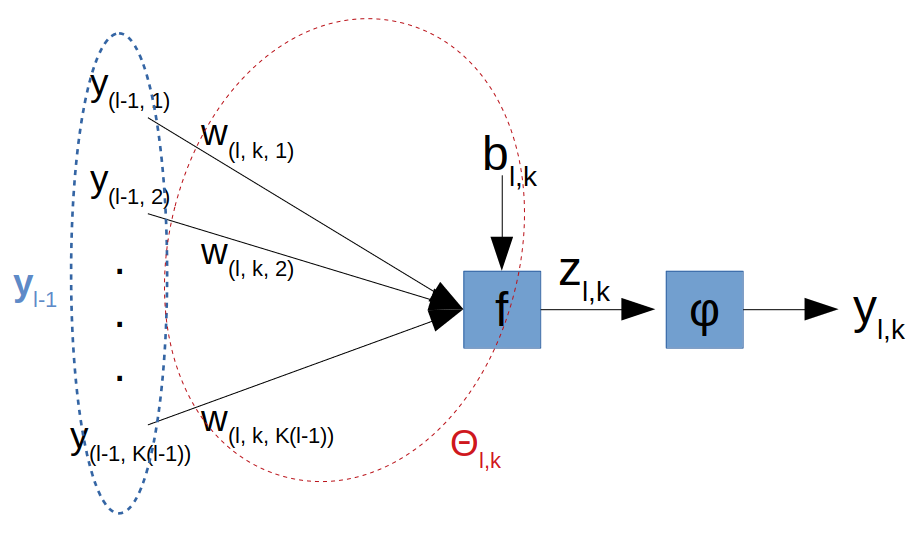}
    \caption{Representation of the $k$-th neuron of the $l$-th layer of an ANN. The input $\mathbf{y}_{l-1}$ is weighted by the parameters $\Theta_{l,k}$, passes through some affine function $f(\cdot)$ producing the post-synaptic potential $z_{l, k}$ which is fed to the activation function $\varphi(\cdot)$, producing the output $y_{l, k}$}
    \label{fig::notation}
\end{figure}
\end{center}
In this section we introduce the notation to be used in the rest of this work. Let us assume we work with an acyclic, multi-layer artificial neural network composed of $N$ layers, where layer $l=0$ is the input layer and $l=N$ the output layer. The ensemble of all the trained parameters in the ANN will be indicated as $\mathbf{\Theta}$. Each of the $l$ layers is made of $K_l$ neurons (or filters for convolutional layers). Hence, the $k$-th neuron ($k\in [1, K_l]$) in the $l$-th layer has:
\begin{itemize}
    \item $y_{l, k}$ as its own output.
    \item $\mathbf{y}_{l-1}$ as input vector.
    \item $\Theta_{l, k}$ as its own parameters: $\mathbf{w}_{l, k}$ are the weights (from which we identify the $j$-th as $w_{l, k, j}$) and $b_{l, k}$ is the bias.
\end{itemize}
Each of the neurons has its own activation function $\varphi_{l, k}(\cdot )$ to be applied after some affine function $f_{l, k}(\cdot)$ which can be convolution, dot product, adding residual blocks, batch normalization or any combination of them. Hence, the output of a neuron can be expressed by
\begin{equation}
    \label{eq::defy}
    y_{l, k} = \varphi_{l, k}\left[ f_{l, k} \left(\theta_{l, k}, \mathbf{y}_{l-1} \right) \right]
\end{equation}
We can simplify \eqref{eq::defy} if we define the \emph{post-synaptic potential} (or equivalently, the pre-activation potential) $z_{l, k}$ as
\begin{equation}
    \label{eq::defz}
    z_{l, k} = f_{l, k} \left(\theta_{l, k}, \mathbf{y}_{l-1}\right)
\end{equation}
As we are going to see, the post-synaptic potential will be central in our method and analysis and encloses the true essence of the proposed regularization strategy. A summary of the introduced notation is graphically represented in Fig.~\ref{fig::notation}.

\subsection{Effect of weight decay on the post-synaptic potential}
\label{ssec::L2}
Most of the learning strategies use the well-known $\ell_2$ regularization term
\begin{equation}
    \label{eq::L2regu}
    R_{\ell_2}(\mathbf{\Theta}) = \frac{1}{2}\sum_{l}\sum_{k}\sum_{j} \theta_{l, k, j}^2
\end{equation}
Eq.~\eqref{eq::L2regu} is minimized together with the loss function $L(\cdot)$; hence, the overall minimized function is
\begin{equation}
    \label{eq::J}
    J(\mathbf{\Theta}, \hat{\mathbf{y}}) = \eta L(\mathbf{\Theta}, \hat{\mathbf{y}}) + \lambda R_{\ell_2}(\mathbf{\Theta})
\end{equation}
where $\hat{\mathbf{y}}$ is the desired output and $\eta, \lambda$ are positive real numbers, and commonly in range $(0, 1)$. All the update contributions are computed using the standard back-propagation strategy. Let us focus here, for sake of simplicity, on the regularization term \eqref{eq::L2regu}. Minimizing it corresponds to adopt the commonly named \emph{weight decay} strategy for which we have the following update rule:
\begin{equation}
    \label{eq::wd}
    \theta_{l, k, j}^{t+1} := (1 - \lambda) \theta_{l, k, j}^{t} 
\end{equation}
This generates a perturbation of the output for the corresponding neuron resulting in a perturbation of the post-synaptic potential:
\begin{equation}
    \label{eq::variation}
    \Delta z_{l, k} = z_{l,k}^{t+1} - z_{l,k}^{t}
\end{equation}
How does the $\ell_2$ regularization affect $z_{l, k}$?\\
Clearly, minimizing \eqref{eq::L2regu} means that $\theta_{l, k, j} \rightarrow 0 \forall l, k, j$. Now, if we have an input pattern for our network $\mathbf{y}_0$, it is straightforward, according to \eqref{eq::defz} and \eqref{eq::wd}, that $z_{l, k} \rightarrow 0$, as all the parameters will be zero.\\
Under this assumption, we can say that the weight decay strategy implicitly aims to focus on peculiar regions of the mostly-used activation functions: in the case we use \emph{sigmoid} or \emph{hyperbolic tangent}, we have the maximum value for the derivative for $z_{l, k}\approx 0$; while for the ReLU activation we are close to the function discontinuity. Is this the real essence of weight decay and one of the reasons it helps in the generalization?\\
Starting from these very simple observation, we are now going to formulate a regularization term which explicitly minimizes $z_{l, k}$.

\subsection{Post-synaptic regularization}
In the previous section we have observed that, in the typical deep learning scenario, weight decay minimizes the post-synaptic potential, focusing the output of the neuron around some particular regions, which might help in the signal back-propagation and, indirectly, favor the generalization.\\
If we wish to explicitly drive the output $y_{l, k}$ of the neuron, or better, its post-synaptic potential, we can impose an $\ell_2$ regularization on $z_{l,k}$:
\begin{equation}
    \label{eq::Rz}
    R = \frac{1}{2}\sum_{l}\sum_{k} \left(z_{l, k}\right)^2
\end{equation}
where $k$ is an index ranging for all the neurons in the $l$-th layer. We can split \eqref{eq::Rz} for each of the $k$ neurons in the $l$-th layer:
\begin{equation}
    \label{eq::Rzlbyl}
    R_{l, k} = \frac{1}{2} \left(z_{l, k}\right)^2
\end{equation}
In case we desire to apply the regularization \eqref{eq::Rzlbyl}, we can use the chain rule to check what is the update felt by the parameters of our model:
\begin{equation}
    \label{eq::dwz1}
    \frac{\partial R_{l, k}}{\partial \theta_{l,k,j}} = \frac{\partial R_{l, k}}{\partial z_{l, k}} \cdot \frac{\partial z_{l, k}}{\partial \theta_{l,k,j}} = z_{l, k} \cdot \frac{\partial z_{l, k}}{\partial \theta_{l,k,j}}
\end{equation}
Expanding \eqref{eq::dwz1}, we have
\begin{equation}
    \label{eq::rgenup}
    \frac{\partial R_{l, z}}{\partial \theta_{l,k,j}} =  \frac{\partial z_{l, k}}{\partial \theta_{l,k,j}} \cdot \left( b_{l, k} + \sum_i w_{l, k, i} y_{\left(l-1, i\right)}\right)
\end{equation}
Here we need to differentiate between bias and weight cases: if $\theta_{l,k,j}$ is the bias then \eqref{eq::rgenup} can be easily written as
\begin{equation}
    \label{eq::rgbiasup}
    \frac{\partial R_{l, k}}{\partial b_{l,k}} = b_{l,k} + \sum_i w_{l, k, i} y_{\left(l-1, i\right)}
\end{equation}
while, if $\theta_{l,k,j}$ is one of the weights,
\begin{align}
    \frac{\partial R_{l, k}}{\partial w_{l, k, j}} &=  \frac{\partial z_{l,k}}{\partial w_{l, k, j}} \cdot\left[ w_{l, k, j} \frac{\partial z_{l,k}}{\partial w_{k, j}} + \left( b_{l, k} + \sum_{i\neq j} w_{l, k, i} y_{\left(l-1, i\right)}\right)\right]\nonumber\\
    &= w_{l, k, j} \left(\frac{\partial z_{l,k}}{\partial w_{l, k, j}}\right)^2 +\frac{\partial z_{l,k}}{\partial w_{l, k, j}} \left( b_{l,k} + \sum_{i\neq j} w_{l, k, i} y_{\left(l-1, i\right)}\right)\label{eq::22}
\end{align}
From \eqref{eq::22} it is possible to recover the usual weight decay assuming $\frac{\partial z_{l,k}}{\partial w_{l,k, j}} = 1 \forall l,k,j$ and completely neglecting the contribution coming from the other parameters for the same neuron.\\
The variation in the parameter value, according to \eqref{eq::dwz1}, is
\begin{equation}
    \label{eq::updatelz}
    \Delta \theta_{l,k, j} = -\lambda z_{l,k}\frac{\partial z_{l,k}}{\partial \theta_{l,k,j}}
\end{equation}
where $\lambda \in (0, 1)$ as usual. As we are minimizing \eqref{eq::Rz}, we can say that $z_{l,k}$ is a bounded term. Furthermore, looking at $y_{\left(l-1, j\right)}$, we need to distinguish two cases:
\begin{itemize}
    \item $l = 1$: in this case, $y_{\left(0, j \right)}$ represents the input, which we impose to be a bounded quantity.
    \item $l \neq 1$: here we should recall that $y_{\left(l-1, j\right)}$ is the output of the $(l-1)$-th layer: if we minimize the post-synaptic potentials also in those layers, for the commonly-used activation functions, we guarantee it to be a bounded quantity.
\end{itemize}
Hence, as product of bounded quantities, also \eqref{eq::updatelz} is a limited quantity.\\
However, what we aim to minimize is not \eqref{eq::Rzlbyl}, but the whole summation in \eqref{eq::Rz}. If we explicitly wish to write what the regularization contribution to the parameter $\theta_{l, k, j}$ is, we have
\begin{align}
    \frac{\partial R_{p, h}}{\partial \theta_{l, k, j}} = z_{p, h} \cdot \frac{\partial z_{p, h}}{\partial \theta_{l, k, j}}
\end{align}
Here, three different cases can be analyzed:
\begin{itemize}
    \item $p < l$: in this case, the gradient term is $ \frac{\partial z_{p, h}}{\partial \theta_{l, k, j}} = 0$ and the entire contribution is zero.
    \item $p = l$: here, the gradient term $ \frac{\partial z_{p, h}}{\partial \theta_{l, k, j}} = y_{\left(l-1, j\right)}$ if $h=k$, zero otherwise.
    \item $p > l$: this is the most interesting case: regularization on the last layers affects all the previous ones, and such a contribution is automatically computed using back-propagation.
\end{itemize}
Hence, in the most general case, the total update contribution resulting from the minimization of \eqref{eq::Rz} on the $j$-th weight belonging to the $k$-th neuron at layer $l$ is indeed
\begin{equation}
    \label{eq::updatecomplete}
    \Delta \theta_{l, k, j} = -\lambda \left[z_{l, k} \frac{\partial z_{l, k}}{\partial \theta_{l, k, j}} + \sum_{p=l+1}^{L} \frac{\partial R_{p}}{\partial \theta_{l, k, j}} \right]
\end{equation}
where
\begin{equation}
    \frac{\partial z_{l, k}}{\partial \theta_{l, k, j}} = \left\{
    \begin{array}{ll}
        1 & if\ \theta_{l, k, j}\ is\ bias\\
        y_{\left(l-1, j\right)} & if\ \theta_{l, k, j}\ is\ weight
    \end{array}
    \right.
\end{equation}
In this section we have proposed a post-synaptic potential regularization which explicitly minimizes $z_{l, k}$ in all the neurons of the ANN. In particular, we have observed that the update term for the single parameter employs a global information coming from forward layers, favoring the regularization. In the next section, results from some simulations in which PSP regularization is tested are shown.

\section{Experiments}
\label{sec::exp}
In this section we show the performance reached by some of the mostly-used ANNs with our post-synaptic potential regularization (PSP) and we compare it to the results obtained with weight decay. We have tested our regularization on three different datasets: MNIST, Fashion-MNIST and CIFAR-10 on LeNet5, ResNet-18~\cite{he2016deep}, MobileNet~v2~\cite{sandler2018mobilenetv2} and All-CNN-C~\cite{springenberg2014striving}. All the simulationsare performed using the standard SGD with CUDA~8 on a Nvidia~Tesla~P-100 GPU. Our regularization has been implemented using PyTorch~1.1.\footnote{All the source code is publicly available at \url{https://github.com/enzotarta/PSP}}

\subsection{Simulations on MNIST}
\begin{center}
\begin{figure}
    \begin{subfigure}[b]{0.5\textwidth}
            \centering
            \includegraphics[width=\textwidth]{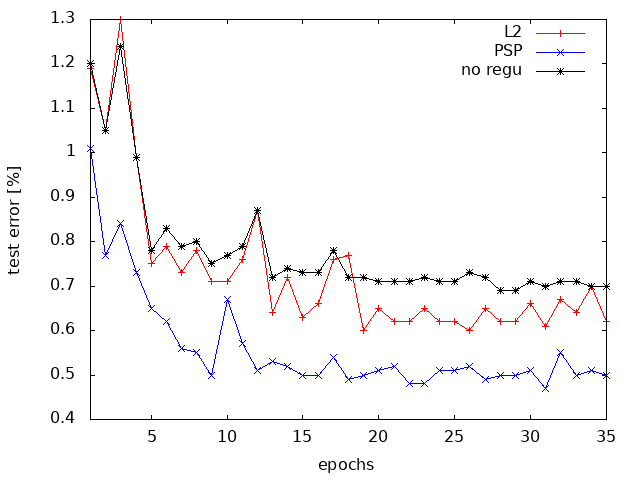}
            \caption{Error on the test set}
            \label{fig::errortestMNIST}
    \end{subfigure}%
    \begin{subfigure}[b]{0.5\textwidth}
            \centering
            \includegraphics[width=\textwidth]{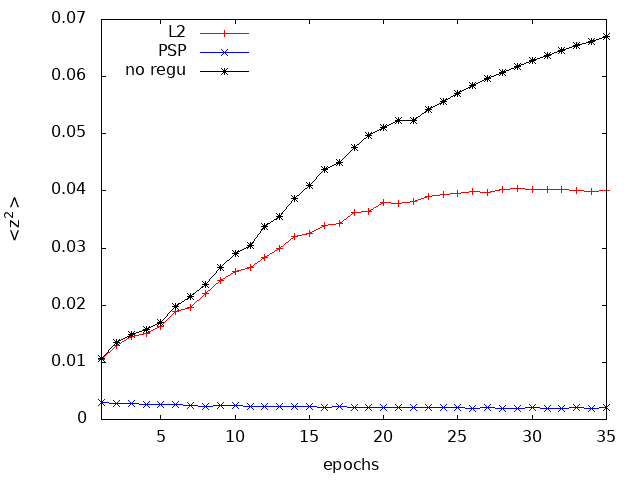}
            \caption{Average of $z^2$ values}
            \label{fig::z2MNIST}
    \end{subfigure}
    \caption{Performance comparison in LeNet5 trained on MNIST (same initialization seed): standard SGD (no regu), $\ell_2$ regularization and post synaptic potential regularization (PSP)}\label{fig:TOF}
\end{figure}
\end{center}

\begin{figure}
    \begin{subfigure}[b]{0.5\textwidth}
            \centering
            \includegraphics[width=\textwidth]{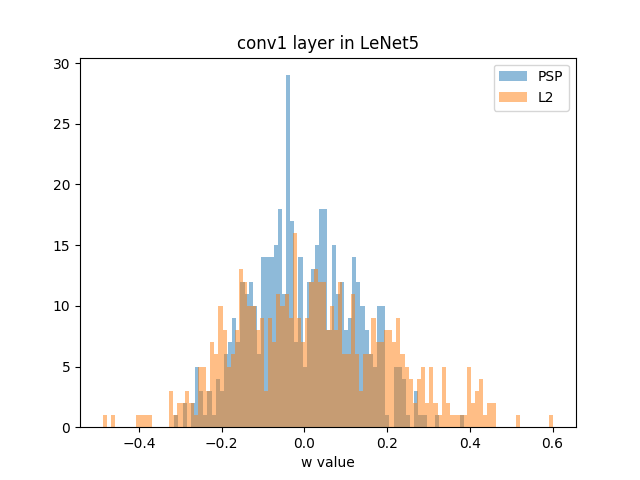}
            \caption{First convolutional layer (conv1)}
            \label{fig::conv1MNIST}
    \end{subfigure}%
    \begin{subfigure}[b]{0.5\textwidth}
            \centering
            \includegraphics[width=\textwidth]{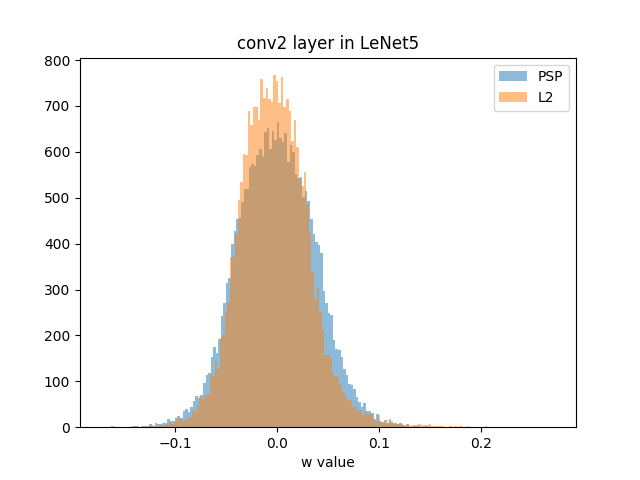}
            \caption{Second convolutional layer (conv2)}
    \end{subfigure}
    
    \vskip\baselineskip
    
    \begin{subfigure}[b]{0.5\textwidth}
            \centering
            \includegraphics[width=\textwidth]{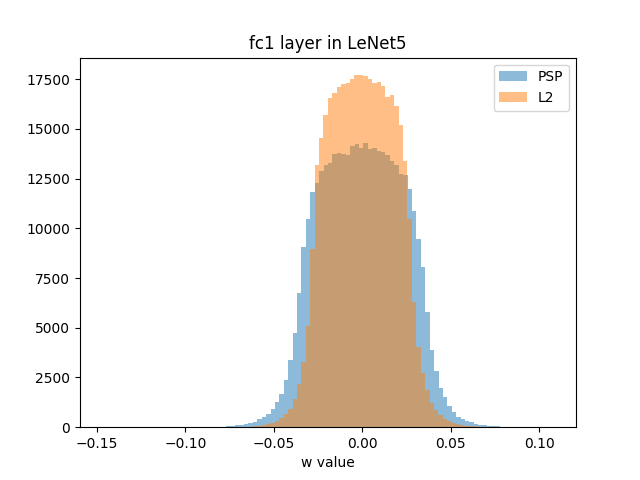}
            \caption{First fullyconnected layer (fc1)}
    \end{subfigure}%
    \begin{subfigure}[b]{0.5\textwidth}
            \centering
            \includegraphics[width=\textwidth]{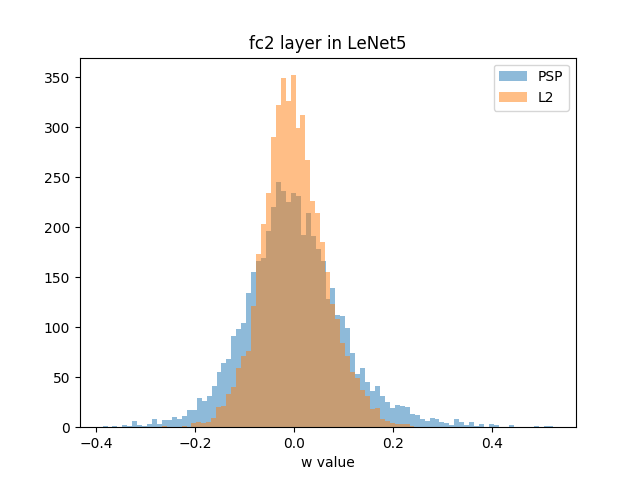}
            \caption{Output layer (fc2)}
    \end{subfigure}
    \caption{Distribution of the parameters in a LeNet5 trained on MNIST with $\ell_2$ regularization ($\ell_2$) and with post synaptic potential regularization (PSP)}
    \label{fig::wdistMNIST}
\end{figure}

As very first experiments, we attempted to train the well-known LeNet-5 model over the standard MNIST dataset~\cite{lecun1998gradient} (60k training images and 10k test images, all the images are 28x28 pixels, grey-scale). We use SGD with a learning parameter $\eta = 0.1$, minibatch size 100. In Fig.~\ref{fig::errortestMNIST}, we show a typically observed scenario in our experimental setting, where we compare standard SGD with no regularization, the effect of $\ell_2$ regularization ($\lambda = 1e-4$) and our pre-activation signal potential regularization (PSP, $\lambda = 0.001$). While the weight decay average performance is 0.64\% (showing improvements from the standard SGD, which is about 0.71\%), PSP final performance is about 0.50\%, with peaks reaching 0.46\%. We would like to emphasize that, to the best of our knowledge, we hit the best ever recorded performance for the current dataset with the same architecture~\cite{chaudhari2016entropy}.\\
We find interesting the behavior of $\left<z^2\right >$ for all the three techniques (Fig.~\ref{fig::z2MNIST}). In the case of standard SGD, the averaged $\left<z^2\right >$ value, as it is not controlled, typically grows until the gradient on the loss will not be zero. For $\ell_2$ regularization, interestingly, it slowly grows until it reaches a final plateau. Finally, in PSP regularization, the $\left<z^2\right >$ value is extremely low, and still slowly decreases. According to the results in Fig.~\ref{fig::errortestMNIST}, this is helping in the generalization.
At this point we can have a further look at what is happening at the level of the distribution of the parameters layer-by-layer. A typical trained parameters distribution for LeNet5, trained on MNIST, is shown in Fig.~\ref{fig::wdistMNIST}. While $\ell_2$ regularization typically shrinks the parameters around zero, PSP regularization does not constrain the parameters with the same strength, while still constrains the pre-activation signal (Fig.~\ref{fig::z2MNIST}). However, contrarily to this, the first convolutional layer, with $\ell_2$ regularization, is less constrained around zero than with PSP (Fig.~\ref{fig::conv1MNIST}). Such a behavior can be explained by \eqref{eq::updatecomplete}: all the regularization contributions coming from all the forward layers (in this case, conv2, fc1 and fc2) affect the parameters in conv1, which are directly conditioning all the $z$ computed in forward layers.

\subsection{Simulations on Fashion-MNIST}
We have decided, as a further step, to test LeNet-5  on a more complex dataset: hence, we have chosen the Fashion-MNIST dataset~\cite{xiao2017online}. It is made of 10 classes of 28x28 grey-scale images representing various pieces of clothing. They are divided in 60k examples for the training set and 10k for the test set. Such a dataset has two main advantages: the problem dimensionality (input, output) is the same as MNIST; hence, the same ANN can be used for both problems, and it is not as trivial as MNIST to solve. 
\begin{table}
    \center
    \caption{LeNet-5 on Fashion-MNIST.}\label{tab1}
    \begin{tabular}{|c|c|}
        \hline
        Technique &  Test set error[\%]\\
        \hline
        SGD+$\ell_2$ &  8.9\\
        SGD+PSP & 8.0\\
        \hline
    \end{tabular}
\end{table}
\noindent Training results are shown in Table~\ref{tab1}. The simulations are performed with $\eta=0.1$, batch size $100$ and $\lambda=0.0001$ for $\ell_2$ regularization while $\lambda=0.001$ for PSP. Here, the difference in the generalization between $\ell_2$ and PSP is wider than the one presented for MNIST: we are able, with the same architecture, to improve the performance by around the 1\% without any other heuristics.

\subsection{Simulations on CIFAR-10}
Moving towards deep architectures, we decided to use CIFAR-10 as dataset. It is made of 32x32 color images (3 channels) divided in 10 classes. The training set is made of 50k images and the test set of 10k samples. This dataset is a good compromise to make the first tests on deep architectures as the training is performed from scratch.\\
Three convolutional architectures have been here tested: MobileNet~v2~\cite{sandler2018mobilenetv2}, ResNet-18~\cite{he2016deep} and All-CNN-C~\cite{springenberg2014striving}. In order to separate the contribution of our regularizer towards other state-of-the-art regularizers, we are going to compare our results with our baseline (same data augmentation, no dropout). All these networks were pre-trained with $\eta=0.1$ for 150 epochs and then learning rate decay policy was applied (drop to 10\% every 100 epochs) for 300 epochs. Minibatch size was set to $128$ and momentum to $0.9$. For standard training, the $\ell_2\ \lambda$ was set to $5e-4$ while for PSP regularization to $0.001$.\\
As we can observe in Table~\ref{tab::cifar}, using PSP-based regularization shows improvements from the baseline and from $\ell_2$ regularization. We speculate that with a proper setting of $\lambda$, PSP can potentially match, or even overcome, top performance marked by state-of-the-art regularizers. 
\begin{table}
    \center
    \caption{Performances on CIFAR-10}\label{tab::cifar}
    \begin{tabular}{|c|c|c|}
        \hline
        Architecture &  Baseline test error[\%] & PSP test error [\%]\\
        \hline
        ResNet-18    & 5.1  &4.6\\
        MobileNet~v2 &7.0    &6.4\\
        All-CNN-C   &9.1    &8.6\\
        \hline
    \end{tabular}
\end{table}

\section{Conclusion}
\label{sec::conclusion}
In this work we have proposed a post-synaptic potential regularizer for supervised learning problems. Starting from the observation that weight decay indirectly shrinks the post-synaptic potential to zero, we have formulated the new PSP regularization. Contrarily to weight decay, it uses a global information coming from other parameters affecting the post-synaptic potential. We have also shown that $\ell_2$ regularization is a special case of our PSP regularization. Looking at the computational complexity, if the autograd~\cite{maclaurin2015autograd} package is used for back-propagation, no significant computational overhead is added.\\
Empirical results show that PSP regularization improves the generalization on both simple and more complex problems, boosting the performance also on deep architectures. Future work includes the application of PSP to recurrent neural networks, tests on networks using non-linear activation functions and the definition of a proper decay policy for PSP regularization.

%
%
%
\bibliographystyle{splncs04}
\bibliography{mybibliography}

\begin{thebibliography}{10}
\providecommand{\url}[1]{\texttt{#1}}
\providecommand{\urlprefix}{URL }
\providecommand{\doi}[1]{https://doi.org/#1}

\bibitem{calimeri2017biomedical}
Calimeri, F., Marzullo, A., Stamile, C., Terracina, G.: Biomedical data
  augmentation using generative adversarial neural networks. In: International
  conference on artificial neural networks. pp. 626--634. Springer (2017)

\bibitem{caruana1997multitask}
Caruana, R.: Multitask learning. Machine learning  \textbf{28}(1),  41--75
  (1997)

\bibitem{chaudhari2016entropy}
Chaudhari, P., Choromanska, A., Soatto, S., LeCun, Y., Baldassi, C., Borgs, C.,
  Chayes, J., Sagun, L., Zecchina, R.: Entropy-sgd: Biasing gradient descent
  into wide valleys. arXiv preprint arXiv:1611.01838  (2016)

\bibitem{cui2015data}
Cui, X., Goel, V., Kingsbury, B.: Data augmentation for deep neural network
  acoustic modeling. IEEE/ACM Transactions on Audio, Speech and Language
  Processing (TASLP)  \textbf{23}(9),  1469--1477 (2015)

\bibitem{elsken2018neural}
Elsken, T., Metzen, J.H., Hutter, F.: Neural architecture search: A survey.
  arXiv preprint arXiv:1808.05377  (2018)

\bibitem{glorot2010understanding}
Glorot, X., Bengio, Y.: Understanding the difficulty of training deep
  feedforward neural networks. In: Proceedings of the thirteenth international
  conference on artificial intelligence and statistics. pp. 249--256 (2010)

\bibitem{goldberg1998regression}
Goldberg, P.W., Williams, C.K., Bishop, C.M.: Regression with input-dependent
  noise: A gaussian process treatment. In: Advances in neural information
  processing systems. pp. 493--499 (1998)

\bibitem{goodfellow2016deep}
Goodfellow, I., Bengio, Y., Courville, A.: Deep learning. MIT press (2016)

\bibitem{gulcehre2016mollifying}
Gulcehre, C., Moczulski, M., Visin, F., Bengio, Y.: Mollifying networks. arXiv
  preprint arXiv:1608.04980  (2016)

\bibitem{he2016deep}
He, K., Zhang, X., Ren, S., Sun, J.: Deep residual learning for image
  recognition. In: Proceedings of the IEEE conference on computer vision and
  pattern recognition. pp. 770--778 (2016)

\bibitem{ioffe2015batch}
Ioffe, S., Szegedy, C.: Batch normalization: Accelerating deep network training
  by reducing internal covariate shift. arXiv preprint arXiv:1502.03167  (2015)

\bibitem{kawaguchi2016deep}
Kawaguchi, K.: Deep learning without poor local minima. In: Advances in neural
  information processing systems. pp. 586--594 (2016)

\bibitem{kingma2014adam}
Kingma, D.P., Ba, J.: Adam: A method for stochastic optimization. arXiv
  preprint arXiv:1412.6980  (2014)

\bibitem{kuko}
Kukacka, J., Golkov, V., Cremers, D.: Regularization for deep learning: {A}
  taxonomy. CoRR  \textbf{abs/1710.10686} (2017),
  \url{http://arxiv.org/abs/1710.10686}

\bibitem{lecun1998gradient}
LeCun, Y., Bottou, L., Bengio, Y., Haffner, P., et~al.: Gradient-based learning
  applied to document recognition. Proceedings of the IEEE  \textbf{86}(11),
  2278--2324 (1998)

\bibitem{lin2017does}
Lin, H.W., Tegmark, M., Rolnick, D.: Why does deep and cheap learning work so
  well? Journal of Statistical Physics  \textbf{168}(6),  1223--1247 (2017)

\bibitem{maclaurin2015autograd}
Maclaurin, D., Duvenaud, D., Adams, R.P.: Autograd: Effortless gradients in
  numpy. In: ICML 2015 AutoML Workshop (2015)

\bibitem{prechelt1998automatic}
Prechelt, L.: Automatic early stopping using cross validation: quantifying the
  criteria. Neural Networks  \textbf{11}(4),  761--767 (1998)

\bibitem{sandler2018mobilenetv2}
Sandler, M., Howard, A., Zhu, M., Zhmoginov, A., Chen, L.C.: Mobilenetv2:
  Inverted residuals and linear bottlenecks. In: Proceedings of the IEEE
  Conference on Computer Vision and Pattern Recognition. pp. 4510--4520 (2018)

\bibitem{simonyan2014very}
Simonyan, K., Zisserman, A.: Very deep convolutional networks for large-scale
  image recognition. arXiv preprint arXiv:1409.1556  (2014)

\bibitem{springenberg2014striving}
Springenberg, J.T., Dosovitskiy, A., Brox, T., Riedmiller, M.: Striving for
  simplicity: The all convolutional net. arXiv preprint arXiv:1412.6806  (2014)

\bibitem{srivastava2014dropout}
Srivastava, N., Hinton, G., Krizhevsky, A., Sutskever, I., Salakhutdinov, R.:
  Dropout: a simple way to prevent neural networks from overfitting. The
  Journal of Machine Learning Research  \textbf{15}(1),  1929--1958 (2014)

\bibitem{tartaglione2018learning}
Tartaglione, E., Leps{\o}y, S., Fiandrotti, A., Francini, G.: Learning sparse
  neural networks via sensitivity-driven regularization. In: Advances in Neural
  Information Processing Systems. pp. 3882--3892 (2018)

\bibitem{tikhonov1943stability}
Tikhonov, A.N.: On the stability of inverse problems. In: Dokl. Akad. Nauk
  SSSR. vol.~39, pp. 195--198 (1943)

\bibitem{wager2013dropout}
Wager, S., Wang, S., Liang, P.S.: Dropout training as adaptive regularization.
  In: Advances in neural information processing systems. pp. 351--359 (2013)

\bibitem{wan2013regularization}
Wan, L., Zeiler, M., Zhang, S., Le~Cun, Y., Fergus, R.: Regularization of
  neural networks using dropconnect. In: International conference on machine
  learning. pp. 1058--1066 (2013)

\bibitem{weigend1991back}
Weigend, A.S., Rumelhart, D.E., Huberman, B.A.: Back-propagation,
  weight-elimination and time series prediction. In: Connectionist models, pp.
  105--116. Elsevier (1991)

\bibitem{xiao2017online}
Xiao, H., Rasul, K., Vollgraf, R.: Fashion-mnist: a novel image dataset for
  benchmarking machine learning algorithms (2017)

\end{thebibliography}

\end{document}